\pdfoutput=1

\documentclass[11pt]{article}

\usepackage[]{acl}

\usepackage{times}
\usepackage{latexsym}

\usepackage[T1]{fontenc}

\usepackage[utf8]{inputenc}

\usepackage{microtype}

\usepackage{amsmath, amscd, amssymb, amsthm}
\usepackage{graphicx}
\usepackage{bm}
\usepackage{bbm}
\usepackage{booktabs}
\usepackage{enumitem}
\usepackage{footnote}
\usepackage{mathtools}
\usepackage{multicol}
\usepackage{multirow}
\usepackage{natbib}
\usepackage{outlines}
\usepackage{pifont}%
\usepackage{subcaption}
\usepackage{soul}
\usepackage{tabularx}
\usepackage{tabulary}
\usepackage{url}
\usepackage{xcolor}
\usepackage[utf8]{inputenc}
\usepackage{hyperref}
\usepackage{cleveref}
\usepackage{diagbox}
\usepackage{tablefootnote}

\newcommand{\norm}[1]{\| #1 \|}

\title{Sentence-T5: Scalable Sentence Encoders \\ from Pre-trained Text-to-Text Models}

\author{Jianmo Ni, Gustavo Hern\'{a}ndez \'{A}brego, Noah Constant, Ji Ma,
\\\textbf{Keith B. Hall, Daniel Cer, Yinfei Yang}\\
  {\rm Google Research}\\Mountain View, CA
}

\begin{document}
\maketitle
\begin{abstract}
We provide the first exploration of sentence embeddings from text-to-text transformers (T5). Sentence embeddings are broadly useful for language processing tasks. While T5 achieves impressive performance on language tasks cast as sequence-to-sequence mapping problems, it is unclear how to produce sentence embeddings from encoder-decoder models.
We investigate three methods for extracting T5 sentence embeddings: two utilize only the T5 encoder and one uses the full T5 encoder-decoder model. To support our investigation, we establish a new sentence representation transfer benchmark, SentGLUE, which extends the SentEval toolkit to nine tasks from the GLUE benchmark~\citep{wang-etal-2018-glue}. Our encoder-only models outperforms Sentence-BERT~\citep{Reimers2019SentenceBERTSE} and SimCSE~\citep{Gao2021SimCSESC} sentence embeddings on both SentEval and SentGLUE transfer tasks, including semantic textual similarity (STS). Scaling up T5 from millions to billions of parameters is found to produce consistent further improvements. Finally, our encoder-decoder method achieves a new state-of-the-art on STS when using sentence embeddings.\footnote{Our models are released at \url{https://tfhub.dev/google/collections/sentence-t5/1}.}

\end{abstract}

\section{Introduction}

Sentence embeddings providing compact meaning representations that are broadly useful for a variety of language processing tasks include classification, question-answering, semantic retrieval, bitext mining, and semantic similarity tasks. 
Sentence embedding models have been trained using a variety of methods including: supervised tasks such as natural language inference~\cite{Conneau2017SupervisedLO,Gao2021SimCSESC} or with semi-structured data such as question-answer pairs~\cite{Cer2018UniversalSE}; translation pairs~\cite{yang-etal-2020-multilingual,fangxiaoyu2020}; paraphrasing pairs \cite{Wieting2016TowardsUP} and adjacent sentence pairs~\cite{Kiros2015SkipThoughtV,quickthought}. Recent work has shown that scaling up model parameters and leveraging pre-trained models~\cite{Devlin2019BERTPO,roberta} are two effective approaches to improve performance~\cite{Reimers2019SentenceBERTSE,reimers-gurevych-2020-making,cmlm,Gao2021SimCSESC}.

\begin{figure}[t]
    \centering
    \includegraphics[width=\linewidth]{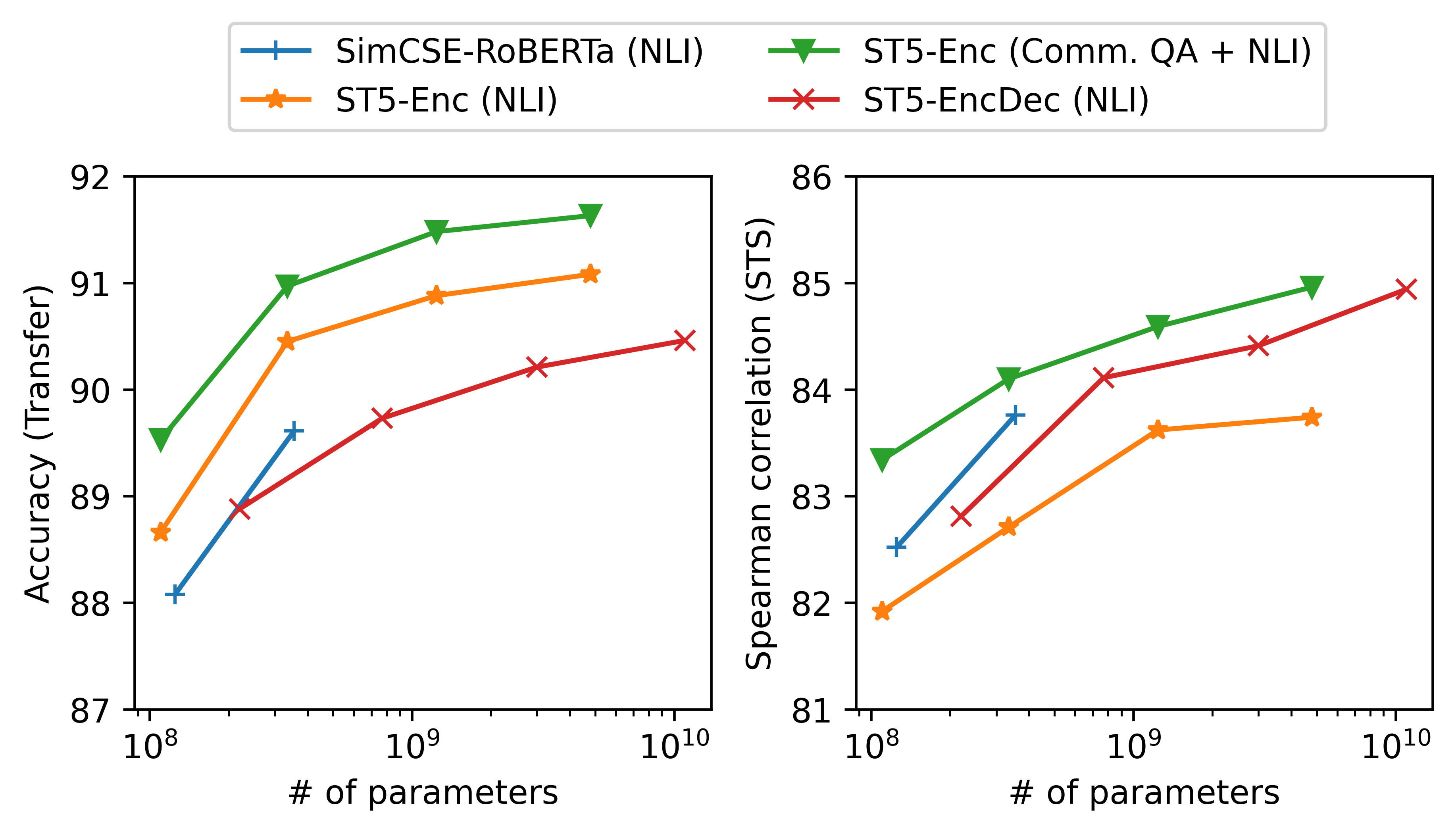}
    \caption{Scaling up our ST5 model size improves performance on SentEval (left) and STS (right).
    }
    \label{fig:scale_up}
\end{figure}

\begin{table}[t]
\small
    \centering
    \resizebox{\linewidth}{!}{%
    \begin{tabular}{l|r|r}
             & Transfer & STS \\ \hline
         ST5-EncDec (11B params)    & 90.46 & 84.94 \\ 
         ST5-Enc   (11B params) 
        & \textbf{91.63} & \textbf{84.96} 
         \\
         \hline
         SimCSE-RoBERTa (large)~\cite{Gao2021SimCSESC} & 90.23\tablefootnote{SimCSE-RoBERTa achieves the best performance on transfer tasks by adding an additional masked language model loss during training while ST5 and other models don't.} & 83.76 \\  %
         SBERT (large)~\cite{Reimers2019SentenceBERTSE} & 87.69  & 76.55 \\
         USE~\cite{Cer2018UniversalSE} & 85.10 & 71.22 \\
         InferSent~\cite{Conneau2017SupervisedLO} & 85.59 & 65.01 \\
         
    \end{tabular}
    } %
    \caption{ST5 versus notable sentence embedding models on SentEval tasks. The reported numbers are the average of transfer tasks and STS tasks}
    \label{tab:exp_summary}
\end{table}

\begin{figure*}
     \centering
     \begin{subfigure}{0.2\textwidth}
     \small
         \centering
         \includegraphics[width=\textwidth]{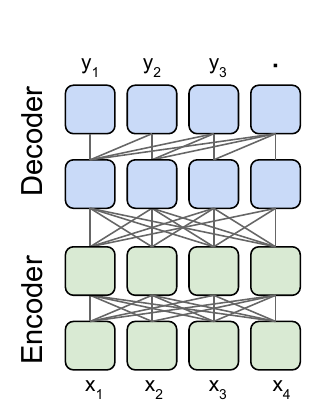}
         \caption{T5 Encoder-Decoder}
         \label{fig:t5}
     \end{subfigure}
     \hspace{2.5em}%
     \begin{subfigure}{0.2\textwidth}
     \small
         \centering
         \includegraphics[width=\textwidth]{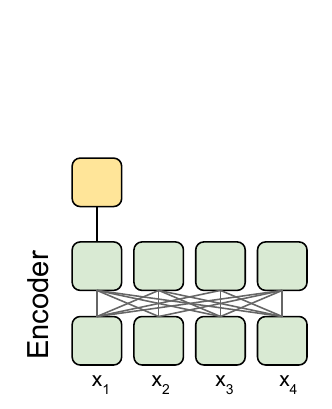}
         \caption{ST5 Encoder-only (ST5-Enc) first}
         \label{fig:enc-first}
     \end{subfigure}
     \hfill
     \hspace{1em}%
     \begin{subfigure}{0.2\textwidth}
     \small
         \centering
         \includegraphics[width=\textwidth]{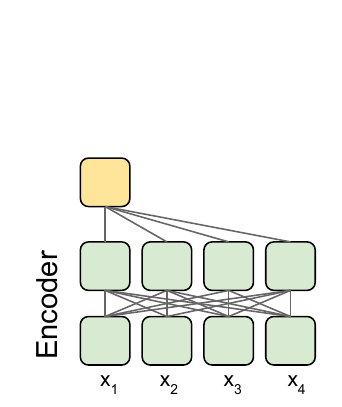}
         \caption{ST5 Encoder-only (ST5-Enc) mean}
         \label{fig:enc-mean}
     \end{subfigure}
     \hfill
     \begin{subfigure}{0.2\textwidth}
     \small
         \centering
         \includegraphics[width=\textwidth]{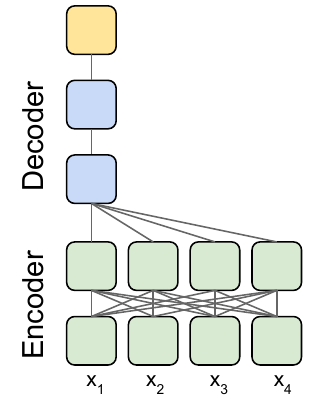}
         \caption{ST5 Encoder-Decoder (ST5-EncDec) first}
         \label{fig:enc-dec-first}
     \end{subfigure}
        \caption{Architecture diagrams for T5 and three ST5 variants to extract sentence representations from T5.}
        \label{fig:st5-arch}
\end{figure*}

We explore sentence embeddings from a new family of pre-trained models: Text-to-Text Transfer Transformer~(T5)~\cite{2020t5}. 
Unlike encoder-only models, which use a transformer encoder to predict random masked tokens, T5 uses an encoder-decoder architecture and a generative span corruption pre-training task. T5 models can be scaled up to hundreds of billions of parameters \citep{Fedus2021SwitchTS} and have achieved state-of-the-art performance on a broad range of NLP tasks including GLUE~\cite{wang-etal-2018-glue} and SuperGLUE~\cite{superglue}. However, it is difficult to efficiently apply T5 to some tasks such as retrieval or clustering. To score retrieval candidates, T5 would need to perform full inference with cross-attention on each query-candidate pair. In contrast, sentence embeddings allow for efficient retrieval and clustering \citep{Gillick2018EndtoEndRI,Reimers2019SentenceBERTSE,yang-etal-2020-multilingual}.

As shown in \cref{fig:st5-arch}, we explore three ways of turning a pre-trained T5 encoder-decoder model into a sentence embedding model: (i)~using the first token representation of the encoder; (ii)~averaging all token representations from the encoder; (iii)~using the first token representation from the decoder.
We evaluate the quality of the resulting sentence embeddings on sentence transfer tasks using Sent\-Eval~\cite{senteval} and on semantic textual similarity~\cite{agirre-etal-2012-semeval,agirre-etal-2013-sem,agirre-etal-2014-semeval,agirre-etal-2015-semeval,agirre-etal-2016-semeval,Cer2017SemEval2017T1}.
We contrast raw re\-presentations from pre-trained T5 models with those learned through fine-tuning on natural language inference (NLI) and Retrieval Question-Answering (ReQA)~\cite{ahmad-etal-2019-reqa} using dual encoders and contrastive learning~\cite{Conneau2017SupervisedLO,Cer2018UniversalSE,yang-etal-2018-learning,Gao2021SimCSESC}. We introduce a multi-stage contrastive learning recipe involving fine-tuning first on ReQA and then on NLI.
Finally, we investigate scaling our T5 sentence embedding model up to 11B parameters.
As shown in \cref{fig:scale_up}, transfer tasks and STS both improve with increased model capacity. 

To our knowledge, we are the first to study using large-scale pre-trained text-to-text models for sentence representation learning and to scale sentence embedding models up to 11 billion parameters. We summarize our contributions as follows: (i) even without fine-tuning, encoder-only ST5 models perform well on sentence transfer tasks, outperforming state-of-the-art fine-tuned models such as SimBERT and SimRoBERTa~\citep{Gao2021SimCSESC}; (ii) encoder-decoder sentence embedding models achieve strong performance on STS, establishing a new state-of-the-art on sentence embedding based STS; (iii)  contrastive learning is effective for fine-tuning sentence encoders from T5-style pre-trained models, particularly using our proposed two-stage contrastive learning approach; (iv) training ST5 longer and with more data using a contrastive loss leads to consistent improvement on both sentence transfer and STS tasks; (v) creating a new sentence representation transfer benchmark `SentGLUE' which extends the sentence evaluation toolkit \citep{senteval} to nine tasks from GLUE \citep{wang-etal-2018-glue} benchmark and evaluating ST5 and other state-of-the-art models on SentGLUE to compare their transfer performance on these challenging tasks. We name our model \textit{Sentence T5 (ST5)}.

\section{Text-to-Text Transfer Transformers~(T5)}
Text-to-Text transfer transformers (T5)~\cite{2020t5} are gaining popularity due to their competitive performance and ease of use in solving a variety of tasks as simple text-to-text mapping problems.
As shown in \cref{fig:t5}, T5 consists of an encoder-decoder transformer model~\cite{vaswani2017} pre-trained on an unsupervised span corruption task. Though T5 has been successfully applied to numerous NLP tasks, how to extract high quality text representations from T5 remains unexplored.

\section{Sentence T5}

\subsection{Model Architecture}
\label{sec:arch}

In this work we explore three strategies to extract sentence representations from T5, as shown in \cref{fig:enc-first,fig:enc-mean,fig:enc-dec-first}:  
\begin{itemize}
    \item \textbf{Encoder-only first} (ST5-Enc first): The encoder output of the first token is taken as the sentence embedding.
    \item \textbf{Encoder-only mean} (ST5-Enc mean): The sentence embedding is defined as the average of the encoder outputs across all input tokens.
    \item \textbf{Encoder-Decoder first} (ST5-EncDec first): The first decoder output is taken as the sentence embedding. To obtain the decoder output, the input text is fed into the encoder, and the standard ``start'' symbol is fed as the first decoder input.
\end{itemize}

The first two are pooling strategies widely used in encoder-only pre-trained models such as BERT\@. 
Unlike BERT models, T5 models do not have a CLS token at the beginning of each sentence. 
For T5 encoder-decoder models, we assume the decoder is aware of the semantics of the entire input sentence when generating its first token prediction; and if so, the first decoder output embeddings (i.e.~input to the softmax layer) might naturally capture the sentence semantics.

\begin{figure}
    \centering
    \includegraphics[width=0.8\linewidth]{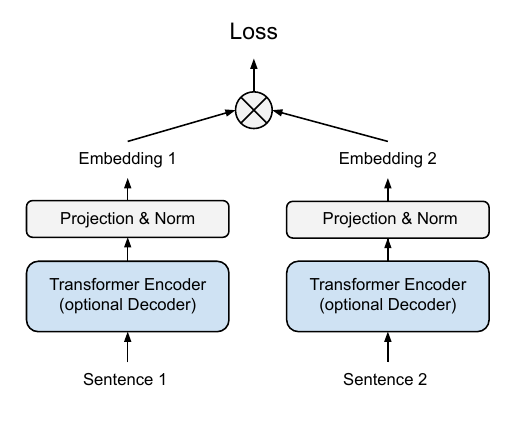}
    \caption{Architecture of the dual encoder model.}
    \label{fig:de}
\end{figure}

For sentence encoder training, 
we adopt a \textit{dual encoder} architecture \citep{Gillick2018EndtoEndRI,Cer2018UniversalSE,Reimers2019SentenceBERTSE}.
As shown in \cref{fig:de}, this architecture consists of two shared-weight transformer modules that encode the inputs.
The transformer module can be either an encoder-only or encoder-decoder architecture.
In our experiments, we initialize the transformer modules from the pre-trained T5 models.
After each module computes a fixed-length representation for its input sentence, a projection layer and L2 normalization are applied to the resulting embeddings.
The projection layer transforms the output to a configurable fixed dimensionality (i.e.~the sentence embedding size).
The embeddings from paired encoding towers can be scored for similarity tasks using a dot-product\footnote{Since L2 normalization is applied to the output of each tower, the dot-product between the embeddings will produce their cosine similarity.} or provide as input to additional layers layers for pairwise classification tasks (e.g., NLI).

\subsection{Contrastive Learning}

Applying contrastive learning to sentence embeddings improves the uniformity of the embeddings space, leading to better performance on downstream tasks such as STS~\citep{Gao2021SimCSESC}. 
We apply contrastive learning to fine-tune the T5 sentence representations.\footnote{In preliminary experiments, we also explored fine-tuning with the classification loss used in InferSent~\cite{Conneau2017SupervisedLO} and Sentence-BERT~\cite{Reimers2019SentenceBERTSE}. However we found fine-tuning for classification on an NLI dataset is inferior to contrastive learning as reported in \citep{Gao2021SimCSESC}}

\subsubsection{Contrastive Loss}
Using a contrastive loss to train a sentence encoder requires paired examples $\mathcal{D}=\{(v_i, v_i^{+})\}$ as a training set, where $v_i$ is an input sentence and $v_i^{+}$ is a related sentence (e.g., that is semantically close).
During training, $v_i^{+}$ is considered as a positive example for $v_i$ and all other examples in the batch are considered as negatives.
The model should learn to pull the positive example closer to the input example while pushing away the negatives.
We operationalize our contrastive loss using an in-batch sampled softmax~\citep{Henderson2017EfficientNL}:
\begin{equation}
    \mathcal{L} = \frac{e^{\text{sim}(v_i, v_i^{+})/ \tau}}{\sum_{j \in \mathcal{B}} { e^{\text{sim}(v_i, v_j^{+}) / \tau} } },
    \label{eq::loss}
\end{equation}
The similarity scoring function is \textit{sim}. $\mathcal{B}$ is a mini-batch of examples and $\tau$ is the softmax temperature. When additional negatives $v_j^{-}$ are provided for input example $v$, the loss can be computed as:
\begin{equation}
    \mathcal{L} = \frac{e^{\text{sim}(v_i, v_i^{+})/ \tau }}{\sum_{j \in \mathcal{B}} { e^{\text{sim}(v_i, v_j^{+})/ \tau}  + e^{\text{sim}(v_i, v_j^{-})/ \tau} }} .
\end{equation}

\subsection{Two-stage Training}
To investigate the effect of additional training data, we explore two-stage training: (i) first training on mined question-answering data from Community QA sites; (ii) then, fine-tune the model on sentence pairs with human annotated NLI labels.

\section{Experimental Setup}

\subsection{Training Corpus}
For two-stage training, we use two datasets: one is collected from web forums while the other is from the Stanford Natural Language Inference (SNLI) dataset~\citep{Bowman2015ALA}. For the first stage and similar to \citet{Cer2018UniversalSE}, we collect 
2 Billion question-answers pairs from community QA websites. During training, the associated answer is considered as the positive example for each input question.
For the second stage, we utilize the contrastive version of the NLI dataset \citep{Gao2021SimCSESC} containing 275K examples, where the positives are the `entailment' hypothesis and premise pairs while the negatives are the `contradict' pairs.

\subsection{Evaluation}
We evaluate using SentEval, which includes 7 transfer and 7 STS tasks~\cite{senteval}. For the transfer tasks, sentence embeddings are evaluated by how well they perform as features for a linear classification model. For STS, embeddings are evaluated by how well their cosine similarities correlate with human annotated similiarity scores.\footnote{Following SimCSE \citep{Gao2021SimCSESC}, we report Spearman's correlation for the `all' setting for all STS tasks which aggregates the data across different subsets.}

\subsection{Configurations}
Our models are implemented using JAX\footnote{\url{https://github.com/google/jax}} and trained on Cloud TPU-v8.
We initialize the dual encoder modules from public T5 checkpoints \footnote{\url{https://github.com/google-research/text-to-text-transfer-transformer}}.
During training, we use Adafactor \citep{Shazeer2018AdafactorAL} as the optimizer and set the learning rate to 0.001. 
Linear decay is applied after 10\% of the total number of training steps, reducing the learning rate to 0 by the end of training.
To fine-tune on NLI we use a batch size of 512, while for the Community QA dataset the batch size is 2048. We use a softmax temperature $\tau$ of 0.01.

\section{Experimental Goals}
Our experiments aim to answer the following:
\begin{itemize}
\item Q1: What is the best way to extract sentence representations from T5?
\item Q2: How well do raw T5 sentence embeddings perform on downstream tasks?
\item Q3: How much do contrastive sentence embedding tasks (e.g., NLI, QA) improve the T5 sentence embeddings.
\item Q4: Can we benefit from scaling up model capacity for better sentence representations?
\end{itemize}

With these goals, we study transfer and STS performance of T5 sentence embeddings using a variety of model and training configurations, comparing ST5 to state-of-the-art methods including SBERT/SRoBERTa \citep{Reimers2019SentenceBERTSE} and SimCSE \citep{Gao2021SimCSESC}.

\begin{table*}[h]
    \small
    \centering
    \resizebox{\linewidth}{!}{%
    \begin{tabular}{l|c|rrrrrrrr}
\textbf{Model} & \textbf{Fine-tune data} & \textbf{MR}    & \textbf{CR}    & \textbf{SUBJ}  & \textbf{MPQA}  & \textbf{SST}   & \textbf{TREC}  & \textbf{MRPC}  & \textbf{Avg}   \\
 \hline
 \rule{-2pt}{10pt}
 BERT (CLS-vector) & N/A & 78.68 & 84.85 & 94.21 & 88.23 & 84.13 & 91.4 & 71.13 & 84.66 \\
 BERT (mean) \textsuperscript{$\clubsuit$} & N/A & 78.66 & 86.25 & 94.37 & 88.66 & 84.40 & 92.80 & 69.45 & 84.94 \\
 ST5-Enc first  & N/A & 76.90 & 86.38 & 90.93 & 88.68 & 80.01 & 94.40 & 66.38 & 83.38 \\
 ST5-Enc mean & N/A & \textbf{86.56} & \textbf{91.31} & \textbf{96.01} & \textbf{90.57} & \textbf{90.77} & \textbf{94.60} & \textbf{72.93} & \textbf{88.96} \\
 ST5-EncDec first & N/A & 79.96 & 77.93	& 91.02 & 84.66 & 86.27 & 84.00 & 68.00 & 81.69 \\
 \hline
 \rule{-2pt}{10pt}
 SBERT-NLI \textsuperscript{$\clubsuit$}         & NLI+MNLI & 83.64 & 89.43 & 94.39 & \underline{89.86} & 88.96 & \underline{89.60} & \underline{76.00} & 87.41 \\
 SimCSE-BERT \textsuperscript{$\clubsuit$}       & NLI & 82.69 & 89.25 & \underline{94.81} & 89.59 & 87.31 & 88.40 & 73.51 & 86.51 \\
 SimCSE-RoBERTa \textsuperscript{$\clubsuit$}    & NLI & \underline{84.92} & \underline{92.00} & 94.11 & 89.82 & \underline{91.27} & 88.80 & 75.65 & \underline{88.08} \\
 \hline
 \rule{-2pt}{10pt}
 ST5-Enc mean & NLI & 86.17 & 91.71 & 94.70 & 90.90 & 90.44 & 90.00 & \textbf{76.70} & 88.66 \\
 ST5-EncDec first & NLI & \textbf{86.22} & 91.60 & 94.05 & 90.93 & 90.72 & 92.60 & 76.06 & 88.88 \\
 ST5-Enc mean & CommQA+NLI & 85.75 & 92.08 & 94.58 & \textbf{90.95} & 91.76 & \textbf{96.40} & 75.19 & 89.53 \\
 ST5-Enc-1.1 mean & CommQA+NLI & 86.12 & \textbf{92.50} & \textbf{94.73} & 90.59 & \textbf{92.15} & 95.80 & 76.52 & \textbf{89.77} \\
    \end{tabular}
    }
    \caption{Performance on transfer tasks on the SentEval benchmark. All models are using the \textbf{Base} architecture. \textsuperscript{$\clubsuit$} results are from \citep{Gao2021SimCSESC}. For all tasks, a logistic regression classifier is trained using the sentence embeddings as features and the classification accuracy on test sets are reported.}
    \label{tab:transfer}
\end{table*}

\begin{table*}[h]
    \small
    \centering
    \resizebox{\linewidth}{!}{%
    \begin{tabular}{l|c|rrrrrrrr}
\textbf{Model} & \textbf{Fine-tune data} & \textbf{STS12}    & \textbf{STS13}    & \textbf{STS14}  & \textbf{STS15}  & \textbf{STS16}   & \textbf{STSb}  & \textbf{SICK-R}  & \textbf{Avg}   \\
 \hline
 \rule{-2pt}{10pt}
 BERT (CLS-vector)                         & N/A & 20.16 & 30.01 & 20.09 & 36.88 & 38.08 & 16.50 & 42.63 & 29.19 \\
 BERT (mean) \textsuperscript{$\clubsuit$} & N/A & \textbf{38.78} & \textbf{57.98} & \textbf{57.98} & 63.15 & 61.06 & 46.35 & 58.40 & 54.81 \\
 ST5-Enc first & N/A & 17.50 & 6.35 & -20.70 & 2.29 & 21.87 & 16.71 & 28.60 & 10.37 \\
 ST5-Enc mean & N/A & 37.78 & 56.83 & 49.37 & \textbf{65.48} & \textbf{64.68} & \textbf{57.51} & \textbf{60.11} & \textbf{55.97} \\
 ST5-EncDec first & N/A & 10.91 & 29.59 & 14.90 & 28.91 & 30.61 & 9.45 & 39.31 & 23.38  \\
 \hline
 \rule{-2pt}{10pt}
 SBERT-NLI \textsuperscript{$\clubsuit$}        & NLI+MNLI & 70.97 & 76.53 & 73.19 & 79.09 & 74.30 & 77.03 & 72.91 & 74.89  \\
 SimCSE-BERT \textsuperscript{$\clubsuit$}      & NLI & 75.30 & 84.67 & 80.19 & 85.40 & 80.82 & 84.25 & 80.39 & 81.57  \\
 SimCSE-RoBERTa \textsuperscript{$\clubsuit$}   & NLI & \underline{76.53} & \underline{85.21} & \underline{80.95} & \underline{86.03} & \underline{82.57} & \underline{85.83} & \underline{80.50} & \underline{82.52}  \\
 \hline
 \rule{-2pt}{10pt}
 ST5-Enc mean & NLI & 77.37 & 83.65 & 80.41 & 86.04 & 81.70 & 84.49 & 79.79 & 81.92 \\
 ST5-EncDec first  & NLI & 77.90 & 85.62 & \textbf{82.24} & 86.81 & 82.13 & 84.98 & 79.97 & 82.81 \\
 ST5-Enc mean & CommQA+NLI & \textbf{78.05} & \textbf{85.84} & 82.19 & \textbf{87.46} & \textbf{84.03} & \textbf{86.04} & 79.75 & \textbf{83.34} \\
 ST5-Enc-1.1 mean & CommQA+NLI & 77.58 & 85.12 & 81.46 & 87.14 & 82.89 & 85.82 & \textbf{80.18} & 82.88 \\
    \end{tabular}
    }
    \caption{Spearman's correlation coefficient ($\times$100) on STS tasks on the SentEval benchmark. All models are using the \textbf{Base} architecture. \textsuperscript{$\clubsuit$} results are from \citep{Gao2021SimCSESC}.}
    \label{tab:sts}
\end{table*}

\section{Results}

Table \ref{tab:transfer} and \ref{tab:sts} provide performance on transfer and STS tasks, respectively.
We compare ST5 models with two types of baselines:
(ii)~a model that extracts sentence embeddings from a pre-trained BERT model, listed in rows 1--2 of each table;
(ii)~the current state-of-the-art sentence embedding models fine-tuned from BERT or RoBERTa, listed in rows 6--8 of each table.

\subsection{Results for Raw T5 Sentence Embeddings}

We first evaluate the T5 sentence embeddings without fine-tuning.
We evaluate all three strategies from section \ref{sec:arch}: (i)~Encoder-only first token, (ii)~Encoder-only mean, and (iii)~Encoder-decoder start token.
For all experiments, we use the encoder or decoder outputs from the T5 transformer directly, without doing any projection. This enables us to fully leverage the embedding capacity from the pre-trained models.

\paragraph{Transfer tasks}
Results for ST5 models using raw embeddings on transfer tasks are shown in rows 3--5 of \cref{tab:transfer}.
Unlike BERT, T5's first token (either for encoder or decoder) is not reserved for a special placeholder (i.e.~CLS) and there are no specific pre-training tasks using the first token's embeddings. Therefore, it is unlikely that without additional fine-tuning the first token's representation would capture the semantics of the whole sentence. Indeed, our experiments show the first token's representation from encoder or decoder are much worse on all SentEval tasks compared to the mean pooling of the encoder-only model.

When mean pooling is applied to the T5's encoder outputs, it greatly outperforms the average embeddings of BERT. 
Notably, even without fine-tuning, 
the average embeddings of the T5's encoder-only outputs outperforms SimCSE-RoBERTa, which is fine-tuned on NLI dataset.
This may be due to the fact that T5 is trained on more data. The original T5 models also included downstream tasks (e.g.~GLUE, SuperGLUE) during pre-training, and this multi-task setting may improve transfer performance. However we note that there are only two SentEval tasks (SST and MRPC) included in GLUE while the other five tasks are not.
As shown in \cref{tab:transfer}, we observe significant improvements on the five tasks that are not included.

\paragraph{STS tasks}
In contrast, we observe weak results on STS tasks using raw T5 sentence embeddings as shown in rows 3--5 of \cref{tab:sts}. 
The mean pooling of T5 embeddings achieves an average STS score of 55.97, slightly better than BERT mean pooling but still worse than models fine-tuned on supervised tasks.
This is similar to findings about the \textit{anisotropy} phenomenon of contextual embeddings from other pre-trained language models such as BERT, RoBERTa \citep{Ethayarajh2019HowCA, Gao2021SimCSESC}. Embedding collapse prevents the model from performing well on distance-related metrics.

\subsection{Results for Fine-Tuning T5 Sentence Embeddings}
We next evaluate ST5 models that are fine-tuned on NLI tasks using our contrastive loss, starting from pre-trained T5 models.

Given that mean pooling performs much better than the first token when using encoder only, we opt to discard the first token model when fine-tuning ST5 models. 
The last three rows of \cref{tab:transfer} show that the transfer performance of ST5 models is very consistent across different embedding extracting strategies after fine-tuning. The best fine-tuned model is 0.57 better than the best raw T5 sentence embeddings.

In \cref{tab:sts}, we see that fine-tuning on the NLI dataset significantly improves the STS task performance of ST5 compared to that without fine-tuning.
This supports the claim that contrastive learning is effective to mitigate embedding collapse for T5-style models.

To investigate the impact of additional training data on contrastive learning, we experiment with the ST5 models first trained on Community QA and then fine-tuned on NLI\@. As shown in \cref{tab:sts,tab:transfer}, fine-tuning on an additional dataset brings a large performance boost for both transfer and STS tasks.
This suggests that we may be able to improve sentence embedding quality further through the mining of additional semi-structured data for continued contrastive learning.

To exclude the effect of mixing in downstream tasks, we also trained a ST5 variant based on the T5 1.1 model which only pre-trained on the C4 dataset \citep{2020t5}. As shown on the last row of \cref{tab:transfer} and \cref{tab:sts}, it achieves comparable performance to the original T5 model, outperforming on most tasks but under-performing on STS.

\subsection{Encoder-only vs.~Encoder-decoder}
In this section, we compare the performance of two architectures: encoder-only and encoder-decoder.

\paragraph{Better generalizability for T5's encoder}

In \cref{tab:transfer}, we saw that the encoder-only Base model performs on-par with the encoder-decoder model on transfer tasks. 
When we scale the ST5 model up from Base to Large, 3B and 11B, the encoder-only models' performance on transfer tasks consistently outperforms the encoder-decoder models as shown in \cref{tab:exp_all}.
This shows that building ST5 on top of the T5's encoder gives strong transfer performance.

Recently, \citet{Chung2021RethinkingEC} have shown that larger output embeddings (i.e.~larger embedding size) effectively prevent the encoder from over-specializing to the pre-training task, thus making the encoder's representations more general and more transferable.
We hypothesize that the decoder in the encoder-decoder architecture can improve the generalizability of the encoder's representation in a similar fashion, as the decoder focuses on optimizing for specific tasks.

\paragraph{Effectiveness of the decoder}
In the last two rows of \cref{tab:sts}, we observe that the encoder-decoder architecture outperforms encoder-only models for all STS tasks.
As we scale up the ST5 model, we also observe improvement on STS tasks. 
As shown in \cref{tab:exp_all}, the ST5 encoder-decoder Large model outperforms the state-of-the-art model SimCSE-RoBERTa Large, improving the Spearman's correlation score from 83.76 to 84.11.

One explanation is that the additional parameters from the decoder are helpful for improving textual similarity tasks. 
Another possibility is that the decoder architecture itself helps to improve the sentence embedding quality.
As shown in \cref{fig:enc-dec-first}, the decoder can be considered as an additional attention pooling layer on top of the encoder outputs. As the decoder's weights are lifted from the pre-trained T5 model, the decoder might learn a better way to add attention pooling over the encoder outputs than mean pooling.

\begin{table}[t]
\small
    \centering
    \begin{tabular}{l|r|r|r|r}
          \backslashbox[30mm]{Model}{\# of params} & Base & Large & 3B & 11B \\ \hline
         ST5-Enc & 110M & 335M & 1.24B & 4.8B \\
         ST5-EncDec & 220M & 770M & 3B & 11B \\
    \end{tabular}
    \caption{Number of parameters for different models.}
    \label{tab:param}
\end{table}

\section{Scaling up Sentence T5}
\label{sec:scale}

We leverage the existing checkpoints from large T5 models to study the effect of scaling sentence encoders. 
The parameters of the T5 models are listed in \cref{tab:param}. 
Note however that ST5-EncDec doesn't fully leverage the model parameters; the decoder's learned self-attention is effectively ignored as only the start token is fed into the decoder.

\begin{table*}[t]
    \small
    \centering
    \resizebox{\linewidth}{!}{%
    \begin{tabular}{l|c|rrrrrrrr}
    \toprule
\textbf{Model} & \textbf{Fine-tune data} & \textbf{MR}    & \textbf{CR}    & \textbf{SUBJ}  & \textbf{MPQA}  & \textbf{SST}   & \textbf{TREC}  & \textbf{MRPC}  & \textbf{Avg}   \\
 \midrule
 \rule{-2pt}{10pt}
ST5-Enc mean (Large) & N/A & 89.13 & 92.69 & 97.06 & \textbf{90.70} & 92.92 & 93.60& 73.74 & 89.98 \\
ST5-Enc mean (3B)  & N/A & 90.35 & 92.77 & 97.43 & 90.15 & 93.85 & \textbf{95.60} & 72.70 & 90.41 \\
ST5-Enc mean (11B)  & N/A & \textbf{91.15} & \textbf{93.33} & \textbf{97.55} & 90.20 & \textbf{94.07} & 94.40 & \textbf{74.26} & \textbf{90.71} \\
 \hline
 \rule{-2pt}{10pt}
  SBERT-NLI Large \textsuperscript{$\clubsuit$} & NLI+MNLI  & 84.88 & 90.07 & 94.52 & 90.33 & 90.66 & 87.40 & 75.94 & 87.69 \\
  SimCSE-RoBERTa Large \textsuperscript{$\clubsuit$} & NLI & \underline{88.12} & \underline{92.37} & \underline{95.11} & \underline{90.49} & \underline{92.75} & \underline{91.80} & \underline{76.64} & \underline{89.61} \\
 \hline
 \rule{-2pt}{10pt}
 ST5-Enc mean (Large) & NLI & 88.82 & 93.43 & 95.73 & \textbf{91.75} & 93.08 & 94.00 & 76.35 & 90.45 \\
 ST5-EncDec first (Large) & NLI & 87.63 & 92.85 & 94.32 & 91.37 & 91.98 & 93.00 & 76.99 & 89.73 \\
 ST5-Enc mean (3B) & NLI & 89.92 & 93.27 & 96.19 & 91.54 & 94.18 & 94.20 & 76.87 & 90.88 \\
 ST5-EncDec first (3B) & NLI & 87.83 & 92.85 & 94.75 & 91.01 & 93.14 & 93.60 & 78.26 & 90.21 \\
 ST5-Enc mean (11B) & NLI  & 90.13 & 93.85 & 96.02 & 91.39 & 93.96 & 95.20 & 76.99 & 91.08 \\
 ST5-EncDec first (11B) & NLI  & 90.00 & 93.94 & 95.01 & 91.53 & 93.85 & 92.20 & 76.70 & 90.46 \\
 ST5-Enc mean (Large) & CommQA+NLI & 88.89 & 93.46 & 95.38 & 91.50 & 94.23 & 96.20 & 77.10 & 90.97 \\
 ST5-Enc mean (3B) & CommQA+NLI & 89.94 & 94.09 & 95.85 & 91.58 & 94.84 & \textbf{96.20} & 77.86 & 91.48 \\
 ST5-Enc mean (11B) & CommQA+NLI & \textbf{90.83} & \textbf{94.44} & \textbf{96.33} & 91.68 & \textbf{94.84} & 95.40 & \textbf{77.91} & \textbf{91.63} \\
 \midrule
 \midrule
 \textbf{Model} & \textbf{Fine-tune data} & \textbf{STS12}    & \textbf{STS13}    & \textbf{STS14}  & \textbf{STS15}  & \textbf{STS16}   & \textbf{STSb}  & \textbf{SICK-R}  & \textbf{Avg} \\
  \midrule
  ST5-Enc mean (Large) & N/A  & 28.01  & 52.60          & 41.35  & 61.28  & 63.58 & 56.31  & 59.48  & 51.80 \\
  ST5-Enc mean (3B) & N/A   & 24.89    & 51.49          & 41.09  & 61.37   & 64.51   & 52.57    & 59.99  & 50.85 \\
 ST5-Enc mean (11B) & N/A   & \textbf{34.97}  & \textbf{60.19}  & \textbf{47.59}  & \textbf{66.40}    & \textbf{70.62}  & \textbf{62.83}  & \textbf{63.57}  & \textbf{58.02}  \\
  \hline
  \rule{-2pt}{10pt}
  SBERT-NLI Large  \textsuperscript{$\clubsuit$}  & NLI+MNLI & 72.27  & 78.46 & 74.90  & 80.99  & 76.25   & 79.23   & 73.75   & 76.55  \\
  SimCSE-RoBERTa Large \textsuperscript{$\clubsuit$}  & NLI & \underline{77.46}  & \underline{87.27}  & \underline{82.36}  & \underline{86.66}  & \underline{83.93} & \underline{86.70}   & \underline{81.95} & \underline{83.76}  \\
\hline
\rule{-2pt}{10pt}
ST5-Enc mean (Large) & NLI & 76.52 & 85.75 & 81.01 & 87.13 & 83.26 & 85.45	& 79.85 & 82.71 \\
ST5-EncDec first (Large) & NLI & 79.15 & 87.42 & 83.61 & 87.64 & 83.92 & 86.35 & 80.64 & 84.11 \\
 ST5-Enc mean (3B) & NLI & 77.13   & 86.73 & 82.53 & 87.36 & 84.51  & 85.71 & \textbf{81.39} & 83.62 \\
 ST5-EncDec first (3B) & NLI & 79.24 & 87.80 & 83.95 & 87.75 & 84.60 & 86.62 & 80.91 & 84.41 \\
 ST5-Enc mean (11B)  & NLI  & 77.42  & 87.50  & 82.51 & 87.47  & 84.88 & 85.61 & 80.77 & 83.74  \\
 ST5-EncDec first (11B) & NLI   & \textbf{80.11} & 88.78 & 84.33 & 88.36 & \textbf{85.55} & \textbf{86.82} & 80.60 & 84.94 \\
 ST5-Enc mean (Large) & CommQA+NLI & 79.10 & 87.32 & 83.17 & 88.27 & 84.36 & 86.73 & 79.84 & 84.11 \\
 ST5-Enc mean (3B) & CommQA+NLI & 79.02   & \textbf{88.80} & 84.33 & \textbf{88.89}   & 85.31 & 86.25 & 79.51 & 84.59  \\
 ST5-Enc mean (11B) & CommQA+NLI & 80.10 & 88.75 & \textbf{84.70} & 88.86 & 85.17 & 86.77 & 80.39 & \textbf{84.96}  \\
 \bottomrule
    \end{tabular}
    }
    \caption{Comparisons of models' performance on SentEval benchmark when scaling up model size. \textsuperscript{$\clubsuit$} results are from \citep{Gao2021SimCSESC}.  The first set of results are for the transfer task; the second set are for the similarity task.}
    \label{tab:exp_all}
\end{table*}

\subsection{Effect on Directly Using T5 Embeddings}
As shown in \cref{tab:exp_all}, the performance on the transfer tasks of directly using T5 embeddings consistently improves as T5 scales up. This corroborates that large pre-trained models can improve transfer performance of sentence embeddings.

On the other hand, increasing the model capacity alone is not enough to mitigate the embedding collapse. Even the embeddings from the T5 11B model still do worse on STS tasks than fine-tuned models.
One reason is that the pre-training snap corruption task of T5 does not require the model to avoid anisotropy (e.g., by using a contrastive loss or regularization). 
This highlights the importance of choosing fine-tuning tasks that are aligned to the goal of similarity/retrieval performance.

\subsection{Improving the ST5 Fine-tuning}

As shown in \cref{tab:exp_all}, we find that scaling up model capacity leads to consistently better performance on all downstream tasks.
For the ST5 11B model, the encoder-only model achieves an average score of 91.08 for transfer tasks which is better than 90.45 from the ST5 Large model; while the encoder-decoder model pushes the STS score to 84.94 that also outperforms the ST5 Large model.
This inspires us to explore even larger model sizes to achieve better sentence embedding quality. 

For STS tasks, we observe that the gain from increasing model size from 3B to 11B is smaller than that from Large to 3B. 
This might be due to the fact that the embedding sizes are fixed for all models in our experiments.
One potential exploration is to increase the sentence embedding size for larger models to fully leverage the model capacity.

\begin{table*}[h]
    \small
    \centering
    \resizebox{\linewidth}{!}{%
    \begin{tabular}{l|c|rrrrrrrrrr}
\toprule
Model            & Sent.\ Embed.\ Fine-tuning      & Score & CoLA   & SST-2 & MRPC  & STS-B & QQP                  & MNLI-m               & MNLI-mm              & QNLI  & RTE   \\
 \hline
 \rule{-2pt}{10pt}
InferSent \citep{wang-etal-2018-glue}  & NLI & 66.71 & 8.60   & 83.90 & 76.50 & 80.20 & 81.70                & 67.80                & -                    & 63.50 & 71.50 \\
SBERT (RoBERTa Base) \textsuperscript{$\clubsuit$} & NLI+MNLI  & 73.40 & 21.22  & 90.83 & 73.34 & 74.08 &  80.75                    &       77.21  & 78.13  & 73.92 & 57.76 \\
SBERT (RoBERTa Large) \textsuperscript{$\clubsuit$} & NLI+MNLI & 75.81 & 20.69  & 93.00 & 73.39 & 76.26 & 82.26                &    79.46    &    80.18    & 75.80 & 60.65 \\
SimCSE (RoBERTa Base) \textsuperscript{$\clubsuit$} & NLI &  77.05 & 35.87  & 90.71 & 76.47 & 83.93 & 81.39  & 70.74  & 72.05 & 76.30 & 60.29 \\
SimCSE (RoBERTa Large) \textsuperscript{$\clubsuit$} & NLI & 76.23 & 40.11  & 93.23 & 70.23 & 81.45 & 84.45 & 73.44 & 73.56  & 75.95 & 60.65 \\
 \midrule
ST5 Enc (Base)    & CommQA+NLI  & 76.89 & 22.73  & 91.40 & 76.88 & 86.58 & 84.55                & 69.73                & 70.00                & 79.54 & 61.73 \\
ST5 Enc (Large)  & CommQA+NLI  & 78.52 & 29.46  & 93.92 & 77.26 & 86.07 & 85.32                & 72.20                & 72.44                & 79.64 & \textbf{66.43} \\
ST5 Enc (3B) & CommQA+NLI   & 79.06 & 34.78 & 94.95 & \textbf{78.71} & 85.84 & 85.78                & 72.38                & 73.10                & 79.70 & 66.06 \\
ST5 Enc (11B) & CommQA+NLI  & \textbf{80.07} & \textbf{43.91} & \textbf{95.30} & 78.46 & \textbf{86.54} & \textbf{86.21}                & \textbf{73.46}                & \textbf{74.42}  & \textbf{80.12} & 66.06 \\
ST5 Enc 1.1 (Base) & CommQA+NLI  & 76.63 & 21.59  & 90.60 & 76.66 & 86.34 & 84.53                & 70.40                & 70.76                & 77.92 & 61.01 \\
ST5 Enc-Dec (Base) & NLI  & 76.37 & 17.46  & 90.71 & 76.44 & 85.98 & 82.65                & 70.49                & 70.96                & 76.20 & 63.18 \\
 \midrule
T5 (Base) \citep{2020t5} & -  & 83.40 & 53.84  & 92.68 & 88.92 & 88.02 & 91.56                & 84.24                & 84.57                & 90.48 & 76.28 \\
\bottomrule
    \end{tabular}
}
\caption{Performance on transfer tasks on the Dev set of the GLUE benchmark. \textsuperscript{$\clubsuit$} denotes that the models are released by HuggingFace. T5 (base) is a cross-attention model and other models are embedding based.}
\label{tab:glue_tasks}
\end{table*}

\begin{figure}[t]
    \centering
    \includegraphics[width=0.9\linewidth]{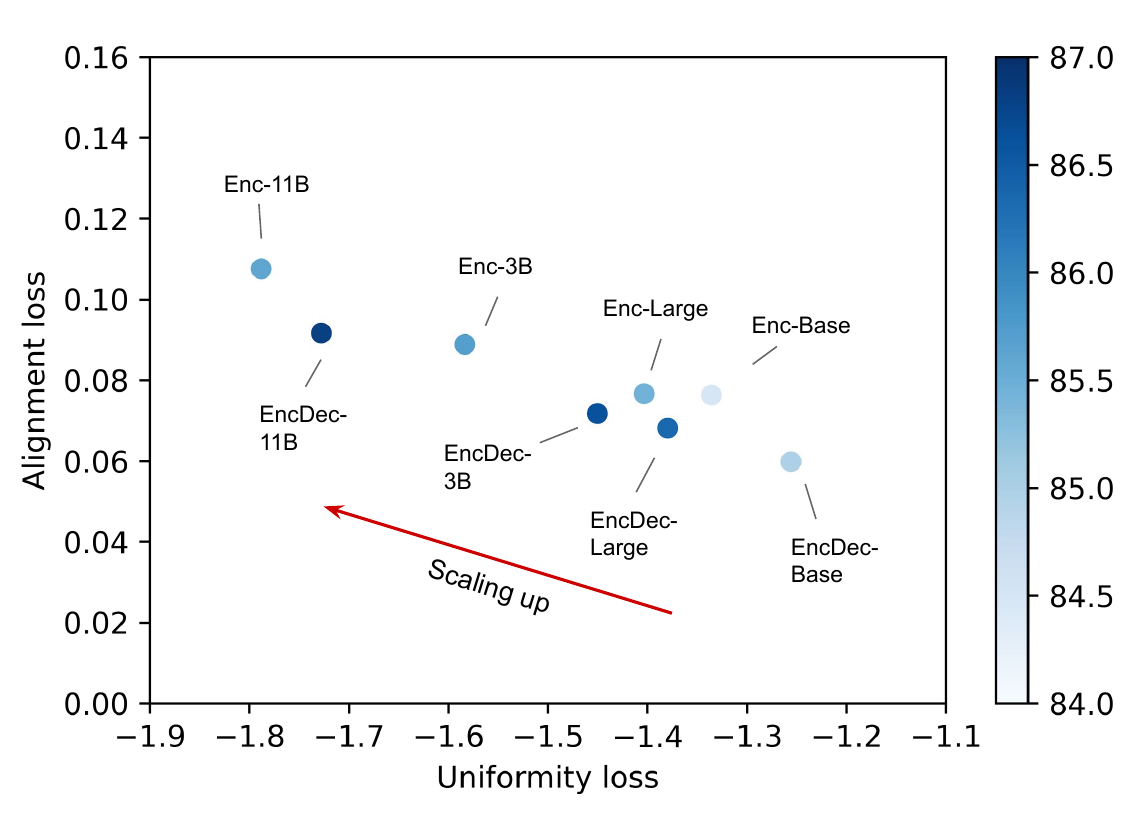}
    \caption{Alignment and uniformity losses for different model sizes. We consider the test split of the STS-B dataset. $\mathcal{L}_\text{align}$ is calculated considering all pairs with score greater than 4. $\mathcal{L}_\text{uniform}$ is computed using all sentences. The colormap denotes the models' Spearman's correlation score.}
    \vspace{-1em}
    \label{fig:align-uniform}
\end{figure}

We further compute the alignment loss and uniformity loss as defined in \citet{Wang2020UnderstandingCR} to measure the quality of the sentence embeddings:
\begin{align}
    \mathcal{L}_\text{align} & = - \displaystyle \mathop{\mathbb{E}}_{v,v^{+} \sim p_{pos}} \norm{f(v)-f(v^{+})} \\
    \mathcal{L}_\text{uniform} & = \log \mathop{\mathbb{E}}_{v,w \overset{\text{i.i.d}}{\sim} p_\text{data}} {e ^ {-2 \norm{f(v) - f(w)}}},
\end{align}
where $p_{pos}$ is all positive data and $p_{data}$ is the data distribution.  $\mathcal{L}_\text{align}$ denotes the expected distance between
embeddings of the positive pairs of data, while $\mathcal{L}_\text{uniform}$ indicates how uniformly the embeddings are distributed.
For both losses, lower numbers indicate better performance.
As shown in \cref{fig:align-uniform}, when models scale up, both the encoder and encoder-decoder models decrease the uniformity loss with only a slight increase in alignment loss.

We seek to investigate whether the effects of larger model size and more training data are additive for better sentence embeddings. 
As shown in the last two rows of \cref{tab:exp_all}, when scaling up to Large and 3B parameters, ST5 further improves on downstream tasks by training on the Community QA dataset in addition to NLI.

\section{SentGLUE Evaluation}

In this section we introduce a new sentence representation transfer benchmark -- SentGLUE -- which extends the sentence evaluation toolkit to nine challenge tasks from GLUE benchmark including: CoLA, SST-2, MRPC, STS-B, QQP, MNLI-m, MNLI-mm, QNLI, RTE~\footnote{We found the WNLI task from GLUE benchmark is too challenge for existing sentence embedding models, thus we exclude it in the current version.}. The GLUE benchmark has been widely adopted for measuring language understanding models.
GLUE tasks are either single sentence or sentence pair classification (e.g. NLI) or similarity (STS) tasks.
The best models on the GLUE leaderboard are fine-tuned cross-attention models like BERT or T5. Such models change all the parameters in the underlying model during fine-tuning and for the pairwise tasks they allow for early fusion of input features from both sentences being compared. For SentGLUE, we introduce the constraint that each input needs to be independently encoded into a fixed embedding space representation that can then be feed to additional layers in order to make a prediction. We believe this best adapts the spirit of the original SentEval benchmark for sentence embeddings to the GLUE benchmark tasks.

From \cref{tab:glue_tasks}, ST5-Enc Base outperforms both SBERT-RoBERTa Base and SimCSE-RoBERTa Base on all SentGLUE tasks except CoLA and MNLI.\footnote{MNLI-m and MNLI-mm experiments for SBERT RoBERTa Large and SimCSE RoBERTa Base are still running at the time of submission.} With its increased model capacity, ST5 Enc 11B's sentence embeddings achieve the best overall performance. Notably, as model size is scaled up, aggregate performance using sentence embeddings approaches that of T5 base. This is remarkable given that T5 base makes use of full cross-attention between sentence pairs and adjusts all of the parameters in the model during fine-tuning.

\section{Conclusion}
In this paper, we study effective methods to build T5 sentence encoders (ST5) from pre-trained models. We propose three architectures and a two-stage contrastive learning method to fine-tune ST5. We compare the difference between encoder-only and encoder-decoder architectures as sentence encoders and analyze their performance on downstream tasks. Through extensive experiments on the Sent\-Eval benchmark, we show that encoder-only models have strong transfer performance while encoder-decoder models perform better on textual similarity tasks. We also demonstrate the effectiveness of scaling up the model size, which greatly improves sentence embedding quality. These findings suggest that future improvements in scale and quality of pre-trained text-to-text models may translate into further gains for sentence encoder models.

\section*{Acknowledgments}
We thank Zora Tung, Daniel Andor, Adam Roberts, Hyung Won Chung, Anselm Levskaya and Livio Baldini Soares for help with the JAX implementation, and Alexis Conneau and Chris Tar for feedback and suggestions.

\bibliographystyle{acl_natbib}
\bibliography{acl2020}

\appendix

\section{Model Inference}
\begin{figure*}[h]
     \centering
     \begin{subfigure}{\textwidth}
     \small
         \centering
         \includegraphics[width=\textwidth, trim=5 0 5 0, clip]{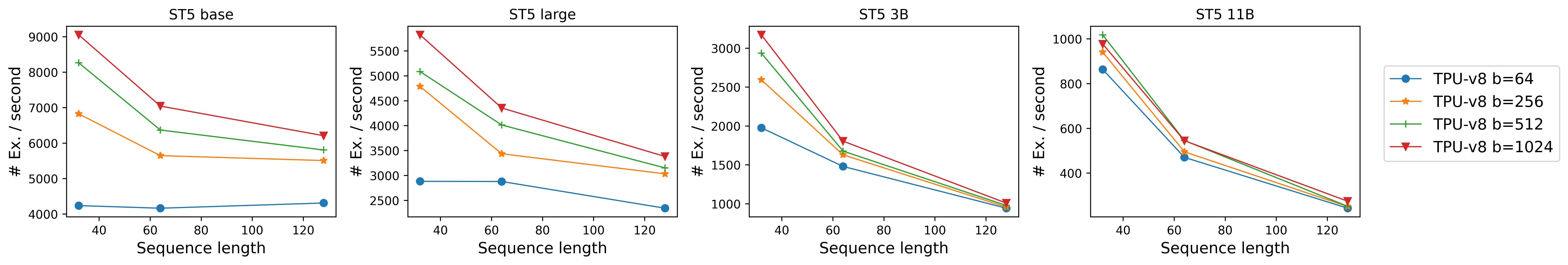}
         \caption{TPU inference speed vs.~sequence length.}
         \label{fig:infer_tpu}
     \end{subfigure} \\
     \begin{subfigure}{\textwidth}
     \small
         \centering
         \includegraphics[width=\textwidth, trim=5 0 5 0, clip]{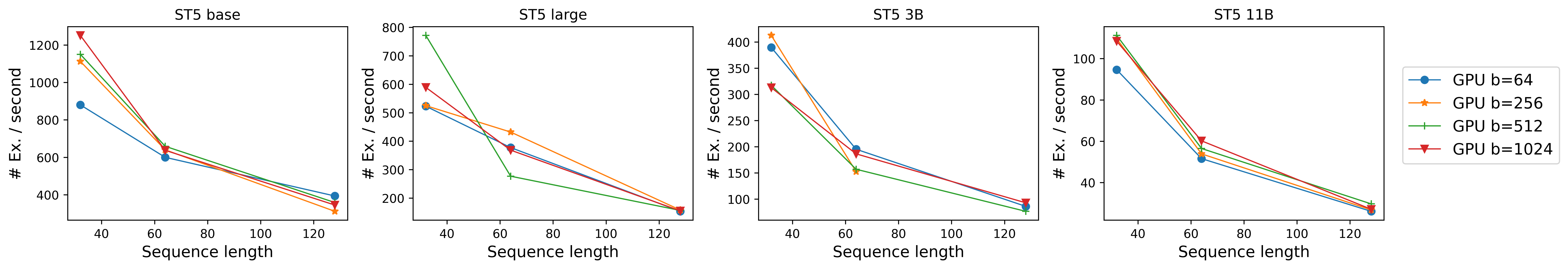}
         \caption{GPU inference speed vs.~sequence length.}
         \label{fig:infer_gpu}
     \end{subfigure} \\
     \begin{subfigure}{\textwidth}
     \small
         \centering
         \includegraphics[width=\textwidth, trim=5 0 5 0, clip]{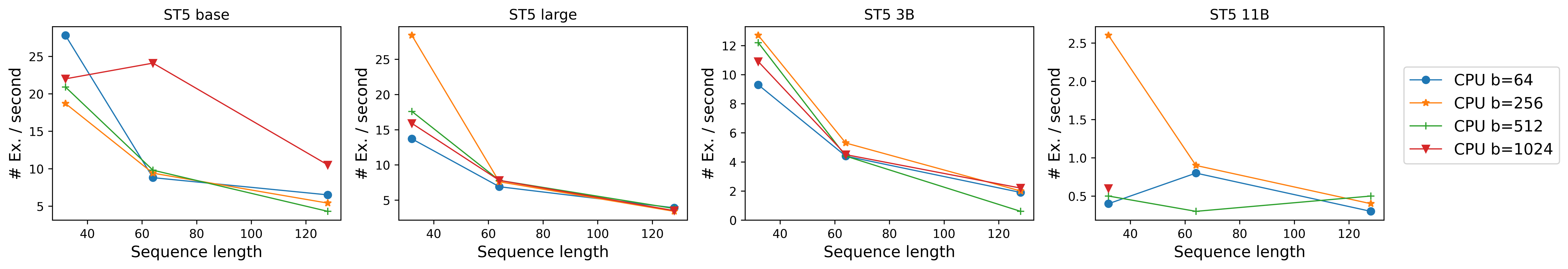}
         \caption{CPU inference speed vs.~sequence length.}
         \label{fig:infer_cpu}
     \end{subfigure}
        \caption{Comparison of inference speed for different model sizes on different platforms.}
        \label{fig:infer}
\end{figure*}

We run ST5 encoder-only on different platforms to investigate the computational cost of inference.
\Cref{fig:infer} summarizes the inference speed for different model sizes, sequence length, batch size and platforms. 
ST5 achieves the fastest inference speed on Cloud TPU-v8.
As we increase the batch size, the inference speed can be further improved.
For the 11B model, we are able to achieve a speed of 274 examples per second for sequence length 128 and batch size 1024.
This shows the feasibility of deploying such large models on TPU hardware.

We also report the speed on Nvidia Tesla V100 GPU and CPU\@. The ST5 11B model is able to run on 4 V100 GPUs with sequence length 128 and batch size 1024, achieving an inference speed of 27 examples per second. For CPU, with batch size 512, ST5 11B achieves 0.5 examples per second.

Although the speed on GPU and CPU are considerably slower than on TPU, the sentence embedding models are much faster than cross-attention based models whose computation time increases quadratically with the number of examples (e.g., clustering 1,000 sentences requires inference over 1 million sentence pairs).

\end{document}